%% file: main.tex
\input{preamble}

\begin{document}

    \maketitle

    \input{section/0-abstract}

    \input{section/1-intro}
    \input{section/2-background}
    \input{section/taxonomy}
    \input{section/3-body}
    \input{section/4-l2p}    
    \input{section/5-discussion}
    \input{section/6-conclusion}

    \bibliography{references}

\input{section/appendix}

\end{document}

%% file: preamble.tex
\pdfoutput=1
\documentclass[11pt]{article}
\usepackage{acl}
\usepackage{times}
\usepackage{latexsym}
\usepackage[T1]{fontenc}
\usepackage[utf8]{inputenc}
\usepackage{microtype}
\usepackage{inconsolata}
\usepackage{graphicx}
\usepackage{enumitem}
\usepackage{tikz}
\usepackage{forest}
\usepackage{adjustbox}
\usetikzlibrary{trees,positioning,shapes,fit,shadows}
\usepackage{wrapfig}

\usepackage{minted}
\usepackage{caption} 
\usepackage{tcolorbox} 
\tcbuselibrary{listingsutf8}
\tcbuselibrary{skins}
\tcbset{
    sharp corners,
    colback = white,
    before skip = 0.2cm,
    after skip = 0.2cm
}  

\usepackage{booktabs}
\usepackage{amssymb}
\usepackage{amsmath}

\newtcolorbox{boxK}{
    sharpish corners,
    boxrule = 0pt,
    toprule = 2pt,
    enhanced,
    colback=red!10, 
    fuzzy shadow = {0pt}{-2pt}{-0.5pt}{0.5pt}{black!35}
}

\newtcolorbox{boxA}{
    sharpish corners,
    boxrule = 0pt,
    toprule = 2pt,
    enhanced,
    colback=blue!10, 
    fuzzy shadow = {0pt}{-2pt}{-0.5pt}{0.5pt}{black!35}
}

\newtcolorbox{boxB}{
    rounded corners,
    boxrule = 0pt,
    toprule = 2pt,
    enhanced,
    colback=blue!20, 
    fuzzy shadow = {0pt}{-2pt}{-0.5pt}{0.5pt}{black!35}
}

\usepackage{listings}
\usepackage{xcolor} 

\definecolor{actioncolor}{rgb}{1,0,0} 
\definecolor{paramcolor}{rgb}{0,0,1}  
\definecolor{predcolor}{rgb}{0,1,1} 

\lstdefinestyle{pddlstyle}{
  language=Python,                
  basicstyle=\ttfamily\small,     
  keywordstyle=\color{blue},      
  commentstyle=\color{gray},      
  stringstyle=\color{purple},     
  morekeywords={action, parameters, precondition, effect, predicates, init, goal, objects}, 
  keywordstyle={\color{red}},     
  emph={parameters, precondition, effect}, 
  emphstyle=\color{blue},         
  showstringspaces=false,         
  frame=single,                   
  breaklines=true,                
  breakatwhitespace=true,         
  escapeinside={(*@}{@*)},         
  aboveskip=12pt,                 
  belowskip=12pt,                  
}

\lstdefinestyle{python-aga}{
language=Python,               
  basicstyle=\scriptsize\ttfamily, 
  numbers=left,                  
  frame=lines,                   
  framesep=2mm,                  
  backgroundcolor=\color{white}, 
  rulecolor=\color{black},       
  breaklines=true,               
  showstringspaces=false,        
  escapeinside={\%*}{*)},        
  keywordstyle=\color{blue},      
  commentstyle=\color{darkgray},     
  stringstyle=\color{red},        
  morekeywords={as, import, from, def, class, return, pass, if, else, for, in, try, except, extract_pddl_action, parse_new_predicates}, 
  moredelim=[s][\color{gray}]{\%\%}{\%\%}, 
  aboveskip=0pt,                 
  belowskip=0pt,                 
  lineskip=0.0mm,                
  xleftmargin=2.5em,
  xrightmargin=1em, 
  linewidth=0.49\textwidth,       
}

\lstdefinestyle{python}{
language=Python,               
  basicstyle=\scriptsize\ttfamily, 
  numbers=left,                  
  frame=lines,                   
  framesep=2mm,                  
  backgroundcolor=\color{white}, 
  rulecolor=\color{black},       
  breaklines=true,               
  showstringspaces=false,        
  escapeinside={\%*}{*)},        
  keywordstyle=\color{blue},      
  commentstyle=\color{darkgray},     
  stringstyle=\color{red},        
  morekeywords={as, import, from, def, class, return, pass, if, else, for, in, try, except, extract_pddl_action, parse_new_predicates}, 
  moredelim=[s][\color{gray}]{\%\%}{\%\%}, 
  aboveskip=0pt,                 
  belowskip=0pt,                 
  lineskip=0.0mm,                
  xleftmargin=2.5em,
  xrightmargin=1em, 
  linewidth=\textwidth,       
}

\usepackage{amssymb}

\title{LLMs as Planning Formalizers: A Survey for Leveraging Large Language Models to Construct Automated Planning Models}

\author{
Marcus Tantakoun$^{1,2}$, {\bf Christian Muise}$^{1,2}$, {\bf Xiaodan Zhu}$^{1,3}$ \\
$^1$Ingenuity Labs Research Institute, Queen's University \\ $^2$School of Computing, Queen's University \\
$^3$Department of Electrical and Computer Engineering, Queen's University\\
\small\texttt{\{20mt1, christian.muise, xiaodan.zhu\}@queensu.ca} \\
}

%% file: section/0-abstract.tex
\begin{abstract}
    Large Language Models (LLMs) excel in various natural language tasks but often struggle with long-horizon planning problems requiring structured reasoning. This limitation has drawn interest in integrating neuro-symbolic approaches within the Automated Planning (AP) and Natural Language Processing (NLP) communities. However, identifying optimal AP deployment frameworks can be daunting and introduces new challenges. This paper aims to provide a timely survey of the current research with an in-depth analysis, positioning LLMs as tools for formalizing and refining planning specifications to support reliable off-the-shelf AP planners. By systematically reviewing the current state of research, we highlight methodologies, and identify critical challenges and future directions, hoping to contribute to the joint research on NLP and Automated Planning.
\end{abstract}

%% file: section/1-intro.tex
\section {Introduction}

    The advent of Large Language Models (LLMs) has marked a significant paradigm shift in AI, sparking claims regarding emergent reasoning capabilities within LLMs \citep{wei2022emergentabilitieslargelanguage} and their potential integration into automated planning for agents \citep{pallagani2023understandingcapabilitieslargelanguage}. While LLMs, due to the prowess of distributed representation and learning, excel at System I tasks, planning---an essential aspect of System II cognition \citep{daniel2017thinking}---remains a significant bottleneck \citep{bengio2020deepsystem2}. Furthermore, LLMs face challenges with long-term planning and reasoning, often producing unreliable plans \citep{valmeekam2024llmscantplanlrms, pallagani2023understandingcapabilitieslargelanguage, momennejad2023evaluatingcognitivemapsplanning}, frequently failing to account for the effects and requirements of actions as they scale \citep{chainofthoughtlessness}, with performance degrading under self-iterative LLM feedback \citep{stechly2023gpt4doesntknowits, valmeekam2023largelanguagemodelsreally, huang2024largelanguagemodelsselfcorrect}.

    \begin{figure}[!htbp]
        \centering
        \includegraphics[width=\linewidth]{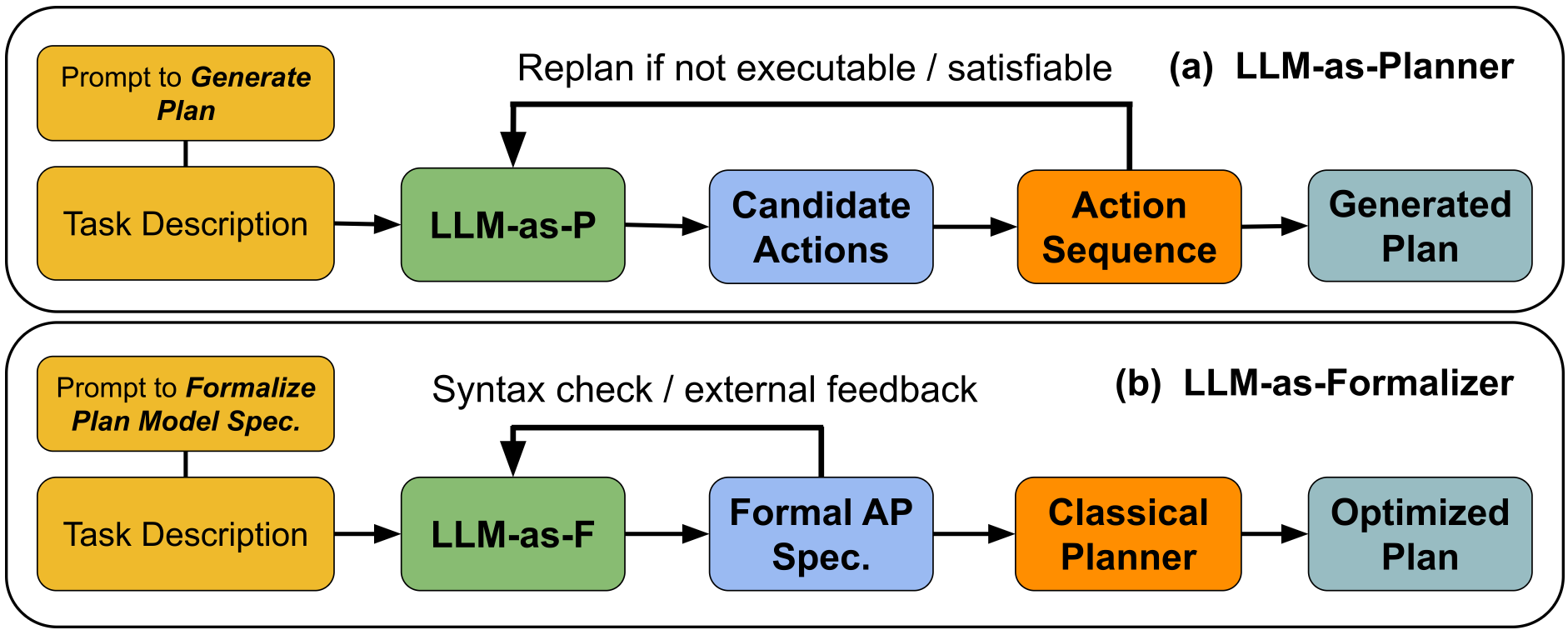}
        \caption{Distinction of planning using LLMs: (a) LLM-as-Planner uses LLMs for direct I/O planning; (b) LLM-as-Formalizer generates planning specifications for existing task planning methods (i.e. PDDL).}
        \label{fig:general-methods}
    \end{figure}
    
    State-of-the-art LLMs have shown limited planning capability by directly generating action sequences---as the correctness, optimality, and reliability of their outputs are not guaranteed.
    Classical Automated Planning (AP) synthesizes plans through structured representation, logic, and search methods which are not subject to the weaknesses mentioned. Meanwhile, LLMs possess promising capabilities at extracting, interpreting, and refining planning model specifications from natural language (NL), acting as complementary components that can enable classical planners to generate robust solutions. Bringing the advantages of AP and LLMs together---LLMs for constructing planning specifications, AP systems for execution---defines the focus of this survey, a paradigm we call \textbf{LLMs-as-Formalizers} for constructing AP models.
    
    This paper is driven by the fragmented landscape in the current literature, where many surveys lack a cohesive overview of LLM integration within this field. Our focus stems from the need to address these gaps and provide a clear framework, highlighting the importance of aligning LLM capabilities with areas where they offer tangible benefits. The motivation for this approach is threefold: (i) \textbf{Planning Accuracy}: LLMs can help ensure that all relevant factors are considered, reducing the risk of overlooked constraints \citep{huang2024planningdarkllmsymbolicplanning}. (ii) \textbf{Adaptability}: LLMs can aid systems in adapting to dynamic environments to capture real-world nuances better, reducing the need for manual edits \citep{dagan2023dynamicplanningllm}. (iii) \textbf{Agnostic Modeling}: LLMs trained on large corpora of diverse data can generalize across various domains without needing task-specific tuning, which reduces the reliance on specialized expertise, a major bottleneck in real-world AP integration \citep{gestrin2024nl2planrobustllmdrivenplanning}. 

    Our survey's taxonomy is divided into three key areas: \textbf{Model Generation}, the process of extracting natural language input---originating from users or the environment---into structured planning model formalizations, further consisting of (i) \textit{Task Generation} (sec. \ref{task-modelling}), which translates initial and goal states along with object assignments; (ii) \textit{Domain Generation} (sec. \ref{domain-modelling}), constructing predicates and action schemas; and (iii) \textit{Hybrid Generation} (sec. \ref{hybrid-modelling}), encapsulating both task instance and domain generation. \textbf{Model Editing} (sec. \ref{model-editing}) systematically addresses code refinement and the repair of errors and inconsistencies in ill-defined planning formalizations. Finally, \textbf{Model Benchmarks} (sec. \ref{model-benchmarks}) encompass assessments of both the performance of LLMs in planning tasks and the quality of LLM-generated planning formalizations.

    \smallskip
    
    To our knowledge, this is the first comprehensive survey of LLM-driven AP-model specification. Our contributions can be summarized as follows:

    \begin{itemize} [leftmargin=3mm, topsep=0mm, itemsep=-2mm]
    
      \item A critical survey of LLM-driven Automated Planning (AP) model generation, editing, and AP-LLM benchmarks, structured within our taxonomy for a comprehensive field overview.
      
      \item A summary of both shared and novel technical approaches for integrating LLMs into AI Planning frameworks alongside  their limitations.
      
      \item We provide insights on key challenges and opportunities,  outlining future research directions for the community. To support future work, we provide \textbf{Language-to-Plan (L2P)}, an open-source Python library that implements landmark papers covered in this survey.

    \end{itemize}

    \noindent We hope this paper will contribute to and facilitate the joint research on Automated Planning and NLP.

%% file: section/2-background.tex
\section{Background}
    
\vspace{-1mm}    \subsection{Automated Planning}
    \vspace{-1mm}
    Automated Planning (AP) focuses on synthesizing action sequences to transition from initial to goal states within its environmental constraints. Its most recognized category is the Classical Planning Problem \citep{russell2003}. This can be formally defined as a tuple \( \mathcal{M} = \langle \mathcal{S}, s\mathcal{^I}, s\mathcal{^G}, \mathcal{A}, \mathcal{T} \rangle\) where \(\mathcal{S}\) is a finite and discrete set of states used to describe the world such that each state \( s \in S\) is defined by the values of a fixed set of variables. \(s\mathcal{^I} \in \mathcal{S}\), \(s\mathcal{^G} \subseteq \mathcal{S}\) represent the initial state and goal world states, respectively. \(\mathcal{A}\) is a set of symbolic actions, and \(\mathcal{T}\) is the underlying transition function which takes the current state \(s^i\) and an action \(a \in \mathcal{A}\) as input and outputs the corresponding next state \(\mathcal{T}(s^i,a) = s^{i+1}\). A solution to a planning problem \(\mathcal{P}\) is a plan $\phi$, which consists of a sequence of actions \(\langle a_1,a_2,...,a_n \rangle\) such that the preconditions of \(a_1\) hold in \(s\mathcal{^I}\), the preconditions of \(a_2\) hold in the state that results from applying \(a_1\), and so on, with the goal conditions all holding in the state that results after applying \(a_n\).

\vspace{0mm}
    \subsection{Planning Domain Definition Language} \label{pddl}
\vspace{0mm}    
    As a fundamental building block of AP, the Planning Domain Definition Language (PDDL) \citep{mcdermottpddl} is one of the most widely used formalisms for encoding planning tasks. These PDDL models serve as formal AP specifications, defining structured symbolic blueprints that enable off-the-shelf external planners to generate robust and optimized solutions. A PDDL model is composed of two files: domain \(\mathbb{DF}\) and problem \(\mathbb{PF}\). \(\mathbb{DF}\) defines the universal aspects of a problem, highlighting the underlying fixed set of rules and constraints. This consists of predicates defining the state space \(\mathcal{S}\), and the set of actions \(\mathcal{A}\). Each \(a \in \mathcal{A}\) is broken down into parameters \textit{Par}\((a)\) defining what types are being used in the action, the preconditions \textit{Pre}\((a)\), and subsequent effects \textit{Eff}\((a)\), encapsulating the transition function \(\mathcal{T}\). \(\mathbb{PF}\)\ consists of a list of objects that ground the domain, the problem's initial states \(s\mathcal{^I}\) and goal conditions \(s\mathcal{^G}\). We provide a concrete example in Appendix \ref{appendix:b}. The standardization and use of PDDL in planning have strongly facilitated sharing and benchmarking. This allows a wide selection of tools to support the validation and refinement of code. Due to its flexibility, clear syntax, and declarative nature, it aligns well with LLMs’ capabilities to translate descriptions into PDDL, as all modern LLMs should have encountered PDDL code in their training corpora.

    \subsection{Large Language Models + Planning} \label{llm-background}

    The advancements of Large Language Models (LLMs) have shown promise in generating highly structured outputs, such as executable code, from NL descriptions \citep{li2023starcoder, planningimprovescodegeneration, nijkamp2023codegenopenlargelanguage}.
    PDDL-LLM research is recent, with initial studies in \citep{miglani2020nltopddl, feng2018extractingactionsequencestexts, simon2021natural, chalvatzaki2023learningreasonscenegraphs}. Currently, researchers are further exploring the nuances of varying pipelines to balance the effectiveness and limitations of LLMs in building such neuro-symbolic frameworks. \citet{huang2024understandingplanningllmagents} survey a limited amount of papers to compose their high-level abstraction of LLM-augmented planning agents. \citet{Pallagani_2024} go beyond the scope of traditional AP, encapsulating broader constructs, whereas \citet{Zhao_2024} provide an extensive overview of LLM-TAMP applications. The work most akin to our paper is \citep{li2024laspsurveyingstateoftheartlarge}, which reviews studies using LLMs in planning with PDDL; however, their survey mainly consists of works comprising \textit{LLMs-as-Planners} (cf. Figure \ref{fig:general-methods}). 

    \begin{boxA}
        \textbf{
        Scope of Survey}: 
        LLMs+AP is an expansive field encompassing many research areas, making it impractical to cover its entirety in a single survey with details and insights. Broadly, LLMs+AP paradigms can be categorized as: (i) \textit{LLMs-as-Heuristics}, where LLMs enhance search efficiency via heuristic guidance \citep{silver2022pddl, simplan, tuisov2025llm, sel2025llms}; (ii) \textit{LLMs-as-Planners}, where they either directly analyze action sequences \citep{DKPROMPT, Lin_2023} or propose plans that are refined through post-hoc methods \citep{gundawar2024robustplanningllmmoduloframework, verifiers-post-hoc, Plansformer, PlanCritic}. In contrast, our survey analyzes (iii) \textit{LLMs-as-Formalizers}, a paradigm in which 
      LLMs are leveraged to construct AP models. Specifically, LLMs assist in defining planning model specifications, supported by domain-independent planners to generate solutions. Our paper surveys approximately 80 existing works, which utilize LLMs to construct planning models, discussing research questions that drive potential directions.

    \end{boxA}

%% file: section/taxonomy.tex
\begin{figure*}[ht]
    \centering
    \begin{minipage}[t]{\textwidth}
        \centering
    \tikzset{
        basic/.style  = {draw, text width=5cm, align=center, font=\sffamily\scriptsize, rectangle, drop shadow=gray!50!black},
        root/.style   = {basic, rounded corners=2pt, thin, align=center, fill=yellow!30, text width=1.5cm, drop shadow=gray!50!black},
        onode/.style  = {basic, thin, rounded corners=2pt, align=center, fill=green!60, text width=1cm, font=\sffamily\small, drop shadow=gray!50!black},
        tnode/.style  = {basic, thin, rounded corners=2pt, align=center, fill=pink!60, text width=1cm, font=\sffamily\tiny, drop shadow=gray!50!black},
        xnode/.style  = {basic, thin, rounded corners=2pt, align=center, fill=blue!20, text width=2cm, font=\sffamily\tiny, drop shadow=gray!50!black},
        wnode/.style  = {basic, thin, rounded corners=2pt, align=left, fill=pink!10!blue!80!red!10, text width=9cm, font=\sffamily\tiny, drop shadow=gray!50!black},
        edge from parent/.style={draw=black, thick}
    }

    \begin{forest} 
        for tree={
            grow=east, 
            parent anchor=east, 
            child anchor=west, 
            edge path={\noexpand\path[\forestoption{edge}] 
                (!u.parent anchor) -- (.child anchor) \forestoption{edge label};},
            s sep=6mm, 
            l sep=8mm  
        }
    [\textbf{LLM\\Model\\Construction} (§\ref{model-construction}), root,  l sep=3mm,
        [\textbf{Model\\Benchmarks\\} (§\ref{model-benchmarks}), xnode,  l sep=16mm,
            [\citet{kambhampati2024llmscantplanhelp};\, \citet{shridhar2021alfworldaligningtextembodied};\, 
            \citet{guan2023leveragingpretrainedlargelanguage};\,
            \citet{xie2024travelplannerbenchmarkrealworldplanning};\, 
            \citet{zheng2024naturalplanbenchmarkingllms};\, 
            \citet{valmeekam2023planbench};\, 
            \citet{acp-bench};\, 
            \citet{stein2024autoplanbenchautomaticallygeneratingbenchmarks};\, \citet{benchmarking-llm};\, 
            \citet{puerta2025roadmap};\, \citet{zuo2024planetariumrigorousbenchmarktranslating};\,
            \citet{text2world}, wnode] ]
        [\textbf{Model\\Editing} (§\ref{model-editing}), xnode,  l sep=16mm,
            [\citet{gragera2023exploring};\, 
            \citet{LLM-error-detection};\, 
            \citet{sikestraversing};\, 
            \citet{caglar2024llmsfixissuesreasoning}, wnode] ]
        [\textbf{Model\\Generation\\} (§\ref{model-generation}), xnode,  l sep=1mm,
            [Hybrid Modeling (§\ref{hybrid-modelling}), tnode, l sep=3mm,
                [\citet{kelly2023there};\, 
                \citet{smirnov2024generatingconsistentpddldomains};\, 
                \citet{zhou2023isrllmiterativeselfrefinedlarge};\, 
                \citet{Sakib_2024};\, 
                \citet{liu2024deltadecomposedefficientlongterm};\, 
                \citet{limits-as-LLM-formalizers};\, 
                \citet{gestrin2024nl2planrobustllmdrivenplanning};\, 
                \citet{dasilva2024methodgeneratecapabilityontologies};\, 
                \citet{LLMFP};\, 
                \citet{ying2023neurosymbolicinverseplanningengine};\,
                \citet{ye2024morpheusmultimodalonearmedrobotassisted};\,
                \citet{han2024interpretinteractivepredicatelearning};\,
                \citet{mahdavi2024leveragingenvironmentinteractionautomated};\,
                \citet{athalye2024predicate};\, 
                \citet{AgentGen}, wnode]]
            [Domain Modeling (§\ref{domain-modelling}), tnode, l sep=3mm,
                [\citet{oates2024using};\, 
                \citet{zhang2024proc2pddlopendomainplanningrepresentations};\,
                \citet{sinha2024leveraging};\,
                \citet{huang2024planningdarkllmsymbolicplanning};\,
                \citet{guan2023leveragingpretrainedlargelanguage};\,
                \citet{llm+al};\, 
                \citet{shah2024autonomously};\,
                \citet{liulearning};\, 
                \citet{wong2023learningadaptiveplanningrepresentations};\,
                \citet{Ding_2023};\, 
                \citet{chen2024languageaugmentedsymbolicplanneropenworld};\,          
                \citet{xie2024makinglargelanguagemodels};\, 
                \citet{sikes2024creating};\,
                \citet{oswald2024largelanguagemodelsplanning}, wnode] ]
            [Task Modeling (§\ref{task-modelling}), tnode, l sep=3mm,
                [\citet{collins2022structuredflexiblerobustbenchmarking};\, 
                \citet{grover2024demonstration};\,
                \citet{lyu2023faithfulchainofthoughtreasoning};\,
                \citet{xie2023translatingnaturallanguageplanning};\, 
                \citet{fast-accurate-llm};\,
                \citet{SafePlanner};\, 
                \citet{singh2024anticipatecollabdatadriven};\, 
                \citet{izquierdo2024plancollabnl};\, 
                \citet{singh2024twostepmultiagenttaskplanning};\, 
                \citet{zhang2024lamma};\,
                \citet{liu2023llmpempoweringlargelanguage};\, 
                \citet{agarwal2024tictranslateinfercompileaccuratetext};\, \citet{agarwal2024llmreasoningplanningsupportingincompleteuser};\, 
                \citet{kalland2024enabling};\,
                \citet{LLM-GROP};\,
                \citet{chen2024autotampautoregressivetaskmotion};\,
                \citet{LGMCTS};\,
                \citet{ViLaIn};\,
                \citet{IALP};\,
                \citet{Birr_2024};\,
                \citet{CaSTL};\,
                \citet{dagan2023dynamicplanningllm};\, \citet{zhang2024pddlegoiterativeplanningtextual};\, 
                \citet{PDDLEGO-dissertation};\,
                \citet{lee2024planning};\,
                \citet{liu2024humanawarenessrobottask};\,
                \citet{NPC-planning};\,
                \citet{paulius2024bootstrappingobjectlevelplanninglarge};\,
                \citet{delarosa2024trippaltravelplanningguarantees};\, \citet{fine2025leveraging}, wnode]] 
            ]
        ]
    \end{forest}
    
        \caption{Taxonomy of research in LLM Planning Model Specification}
        \label{fig:survey-taxonomy}
        \end{minipage}
\end{figure*}

%% file: section/3-body.tex
\section{LLMs for Constructing Automated Planning Models
} \label{model-construction}
We consider the research on leveraging powerful LLMs to assist in constructing planning models to be of critical importance. Verifiable planning modules remain the backbone of planning, ensuring reliability, robustness, and explainability. Note that deploying LLMs themselves in an end-to-end manner to perform planning still falls short of providing soundness guarantees \citep{valmeekam2024planningstrawberryfieldsevaluating} and may have principled weaknesses. We organize the existing works into a taxonomy comprising three key areas: \textbf{Model Generation}, \textbf{Model Editing}, and \textbf{Model Benchmarks}---where the term \textbf{model}, in this context, refers to AP specifications such as PDDL, as illustrated in Figure~\ref{fig:survey-taxonomy}. The tasks require joint efforts from the NLP and automated planning  community.

\subsection{Model Generation} \label{model-generation}
\vspace{-1mm}

A large portion of this survey focuses on \textbf{Model Generation}---extracting and formalizing planning specifications from the user or environment via natural language input. This is further divided into three aspects: \textit{Task Modeling} (sec. \ref{task-modelling}) defines objectives as initial conditions and goal states; \textit{Domain Modeling} (sec. \ref{domain-modelling}) defines the foundational components like entities, actions, and relationships in the system; and \textit{Hybrid Modeling} (sec. \ref{hybrid-modelling}) integrates both aspects to create a complete model, enabling end-to-end planning. A summary of core frameworks is provided in Appendix \ref{appendix:a}.

    To facilitate further discussion on LLM-driven planning model specification, we highlight two key research questions:

\begin{enumerate}[leftmargin=1cm, topsep=1mm, itemsep=0mm]
    \item [\textbf{RQ1:}] How can LLMs accurately align with human goals, ensuring these planning model specifications correctly represent desired expectations and objectives?
    \item [\textbf{RQ2:}] To what extent and granularity of detail can NL instructions be effectively translated into accurate planning model definitions?
\end{enumerate}

\subsubsection{Task Modeling} \label{task-modelling}

For \textit{goal-only specification}, \citet{collins2022structuredflexiblerobustbenchmarking} and \citet{grover2024demonstration} utilize few-shot prompting whereas \textbf{Faithful CoT} \citep{lyu2023faithfulchainofthoughtreasoning} puts heavy emphasis on an interleaving technique of chain-of-thought (CoT) prompting \citep{wei2022chain}. \citet{xie2023translatingnaturallanguageplanning} assess the effectiveness of LLMs in translating tasks with varying levels of ambiguity in both NL and other languages such as Python. \citet{fast-accurate-llm} decompose long-term tasks into sub-goals using LLMs, then executing task planning for each sub-goal with either symbolic methods or MCTS-based LLM planners. \textbf{Safe Planner} \citep{SafePlanner} uses an LLM to convert NL instructions into PDDL goals, enabling a closed-loop VLM-planner to operate based on real-time environmental observations.

Branching to \textit{multi-agent goal collaboration}, \textbf{DaTAPlan} \citep{singh2024anticipatecollabdatadriven} employs an LLM to predict high-level anticipated tasks against human actions, triggering re-planning or new task predictions when deviations occur. \textbf{PlanCollabNL} \citep{izquierdo2024plancollabnl} allocates sub-goals among agents which are then encoded into PDDL, translating LLM output sub-goals into PDDL goals, and modifying the action costs based on LLM recommendations. \textbf{TwoStep} \citep{singh2024twostepmultiagenttaskplanning} decompose multi-agent planning problems into two single-agent problems with mechanisms to ensure smooth coordination. \textbf{LaMMA-P} \citep{zhang2024lamma} uses LLMs to allocate sub-tasks composed of general action sequences, and generates PDDL problem descriptions for each robot's domain. 

Frameworks that handle \textit{complete} PDDL task specifications can be broadly categorized into open-loop and closed-loop approaches. In open-loop systems, \textbf{LLM+P} \citep{liu2023llmpempoweringlargelanguage} uses in-context (IC) examples of a related NL problem and its PDDL representation to generate whole problem files. \textbf{TIC} \citep{agarwal2024tictranslateinfercompileaccuratetext} achieved nearly 100\% accuracy with GPT-3.5 Turbo across LLM+P planning domains by translating the task into intermediate representations, refining them, and processing them through a logical reasoner. \citet{kalland2024enabling} combines a language and an automatic speech recognition model to generate PDDL instances. Recent work converts NL instructions into structured geometric representations that bridge abstract language understanding and spatial reasoning for task and motion planning \citep{LLM-GROP, chen2024autotampautoregressivetaskmotion, LGMCTS}, while other approaches integrate LLMs with Vision Language Models (VLMs) for visual perception to further ground language understanding in spatial contexts \citep{ViLaIn, IALP}.

In closed-loop systems, \textbf{Auto-GPT+P} \citep{Birr_2024} generates the initial state of the problem based on visual perception and an automated error self-correction loop for the generated PDDL goal. \citet{CaSTL} decompose the problem into both PDDL and Python specifications, incorporating a set of constraints into an SMT-based TAMP solver via a Python API. \textbf{LLM+DP} \citep{dagan2023dynamicplanningllm} holds beliefs in uncertain environments to construct possible world states, dynamically updating its internal state and re-plans. \textbf{PDDLEGO} \citep{zhang2024pddlegoiterativeplanningtextual, PDDLEGO-dissertation} performs a recursive task decomposition into sub-goals that enable the agent to gather new observations, progressively refining the problem file until it can develop a solvable plan. In terms of human-in-the-loop collaboration, \citep{lee2024planning} introduce \textbf{PlanAID}, a system that uses Retrieval-Augmented Generation (RAG) to assist LLMs in generating emergency operation plans (EOPs) through improved user interaction. \citet{liu2024humanawarenessrobottask} integrate user information into a hierarchical scene graph of the environment, enabling an LLM to predict human activities and goal states, which are then refined using predicates and domain knowledge to ground problem specifications. \citet{NPC-planning} model NPC behavior by leveraging LLMs conditioned on ``memories'' that represent environmental context.

Rather than extracting PDDL directly, \citet{paulius2024bootstrappingobjectlevelplanninglarge} leverages LLMs to produce Object-Level Plans (OLP), which describe high-level changes to object states and uses them to bootstrap TAMP hierarchically. \textbf{TRIP-PAL} \citep{delarosa2024trippaltravelplanningguarantees} translates intermediate representations via travel points of interest (POI) and user information into dictionaries. \citet{fine2025leveraging} decompose NL goals into predicate-based Python dictionaries, which are then formatted into HDDL decomposition methods. Beyond PDDL, LLM+AP has expanded generating other planning specifications \citep{Logic-LM} such as \textit{temporal logic} (TL) representations \citep{NL2TL}, including TSL \citep{TSL+llm}, STL \citep{NL2STL}, and LTL \citep{nl2spec, lang2LTL, CoT-TL, HTN-LTL}. Recent work shows that LLMs can define the \textit{task search space} by generating successor and goal state Python functions, enabling classical solvers to explore the space efficiently \citep{katz2024thought, a_thought_of_search}.

\begin{boxB}
    \textbf{Summary and Future Directions}: Some methods directly translate NL task descriptions into PDDL \citep{kelly2023there}. Others enhance goal specification by incorporating reasoning chains and few-shot examples \citep{lyu2023faithfulchainofthoughtreasoning, liu2023llmpempoweringlargelanguage}. However, these current approaches rely on an explicit mapping between NL and PDDL code, limiting their processes as code translation tasks. To address ambiguity in minimal task descriptions, future research should develop methods capable of inferring complete and robust PDDL specifications from sparse input, building on prior work that has explored this concept via external perceptual groundings \citep{ViLaIn}, RAG implementations \citep{lee2024planning}, or leveraging LLM commonsense capabilities to capture underlying assumptions and constraints \citep{agarwal2024tictranslateinfercompileaccuratetext}.
\end{boxB}

\subsubsection{Domain Modeling} \label{domain-modelling}

Various works have executed domain modeling in a single query. To better understand cyber-attacks in real-time, \textbf{CLLaMP} \citep{oates2024using} leverages LLMs to extract PDDL action models from Common Vulnerabilities and Exposures descriptions, finding that IC examples are superior to CoT prompting. \citet{zhang2024proc2pddlopendomainplanningrepresentations} introduce \textbf{PROC2PDDL}, which proposes a \textit{Zone of Proximal Development} prompt design—a variant of CoT \citep{vygotsky1978mind}. \citet{sinha2024leveraging} proposes a structured prompt engineering approach to generate domain models in the Hierarchical Planning Definition Language (HPDL). \citet{huang2024planningdarkllmsymbolicplanning} use multiple LLM-generated candidate PDDL action schemas, which are then passed through a sentence encoder to compute the semantic relatedness of code and original NL descriptions. Following the candidate filtering approach, \textbf{pix2pred} \citep{athalye2024predicate} leverages VLMs to propose predicates and determine their truth values in demonstrations.

\citet{guan2023leveragingpretrainedlargelanguage} recognize the impracticality of LLMs generating fully functioning PDDL domains in a single call \citep{kambhampati2024llmscantplanhelp}. Their framework \textbf{LLM+DM (Domain Model)} outlines a generate-test-critique approach \citep{romera2024mathematical, trinh2024solving}, leveraging multiple LLM calls to incrementally build key components of the domain by a dynamically generated predicate list. Similarly, \citet{llm+al} introduce \textbf{LLM+AL}, which uses LLMs to generate action languages in BC+ syntax (extension of Answer Set Programming), while \cite{sikes2024creating} translates Javascript functions to PDDL in incremental stages. \citet{shah2024autonomously} presents \textbf{LAMP}, an extensive series of proposed algorithms that learn abstract PDDL domain models. \textbf{BLADE} \citep{liulearning} bridges language-annotated human demonstrations and primitive action interfaces by tasking an LLM to define the PDDL action model preconditions and effects conditioned on behaviors containing all possible sequences of contact primitive and other behaviors preceding it.

In terms of closed-loop frameworks, \textbf{ADA (Action Domain Acquisition)} \citep{wong2023learningadaptiveplanningrepresentations} tasks LLMs with generating candidate symbolic task decompositions, extracting undefined action names, and iteratively prompting for their definitions. \textbf{COWP} \citep{Ding_2023} handles unforeseen situations in open-world planning by storing the robot's closed-world state when planning fails, triggering a ``Knowledge Acquirer'' module that leverages LLMs to augment action preconditions and effects. Unlike COWP, which relies on predefined error factors, \textbf{LASP} \citep{chen2024languageaugmentedsymbolicplanneropenworld} identifies potential errors from environmental observations, using an LLM to generate error causes in NL, suggesting action preconditions. \citet{xie2024makinglargelanguagemodels} use fine-tuned LLMs for precondition and effect inference from NL actions, and semantic matching to validate actions by comparing inferred preconditions with the current world states.

To evaluate domain quality, \citet{oswald2024largelanguagemodelsplanning} addresses limitations of manual human evaluation \citep{hayton2020narrative, Huang_2012} and string-based comparison methods that assess similarity to ground truth. This study measures equivalence to the ground truth in terms of \textit{operational equivalence}—whether reconstructed domains behave identically to the original by agreeing on the validity of action sequences as plans. To achieve this, the authors decompose ground truth PDDL actions into NL using an LLM, which then tasks them again to reconstruct PDDL domain models for quality assessment.

\begin{boxB}
    \textbf{Summary and Future Directions}: \citet{kambhampati2024llmscantplanhelp, wong2023learningadaptiveplanningrepresentations} use incremental methods that iteratively refine models. Real-world examples have shown to enhance contextual output accuracy \citep{oates2024using, Ding_2023}. The complexity of these frameworks demonstrate that constructing domains are inherently more challenging than task specification. However, by generating and relying on a single domain model, current methods risk rendering the entire planning process invalid if that model fails to capture implicit user constraints. Future approaches should consider generating multiple candidate domains---or specific components, such as predicate definitions---to better accommodate ambiguity and uncertainty in user intent \citep{huang2024planningdarkllmsymbolicplanning, athalye2024predicate}. 
\end{boxB}

\smallskip

\subsubsection{Hybrid Modeling} \label{hybrid-modelling}

Hybrid modeling combines PDDL domain and problem systems. \citet{kelly2023there} extract narrative planning domains and problems from input stories using a one-shot prompt, iterating with a second prompt conditioned on the planner's error message until a successful plan is found. \textbf{ISR-LLM} \citep{zhou2023isrllmiterativeselfrefinedlarge} does not offer any feedback mechanisms to fix PDDL specifications; however, it does introduce self-refinement during the plan generation phase by incorporating the external validator tool, VAL \citep{howey2004val}. \citet{Sakib_2024} generate multiple high-level task plans in Knowledge Graphs (KG), prunes unnecessary components, and feed the task plan to an LLM to extract the PDDL domain and problem files for low-level robot skills. Conversely, \textbf{DELTA} \citep{liu2024deltadecomposedefficientlongterm} initially generates PDDL files and a scene graph, followed by pruning unnecessary details from the graph to focus on relevant items.

\citet{limits-as-LLM-formalizers} further support that LLMs are prone to one-shot generation errors, highlighting the need for intermediate representations before converting to PDDL. \textbf{NL2Plan} \citep{gestrin2024nl2planrobustllmdrivenplanning} is the first domain-agnostic offline end-to-end NL planning system, requiring only minimal description and using pre-processing and automated common sense feedback to interface between the LLM and the user. \citet{smirnov2024generatingconsistentpddldomains} utilize pre-processing steps like JSON markup generation, consistency checks, and error correction loops. Their framework also includes a ``reachability analysis'' pipeline to extract feedback from flawed domains or unreachable problems, alongside a dependency analysis to check predicate usage across both files. \textbf{LLM4CAP} \citep{dasilva2024methodgeneratecapabilityontologies} reduces manual effort, with an LLM-generated ontology being iteratively verified using an LLM to check for syntax errors, hallucinations, and missing elements. \textbf{LLMFP} \citep{LLMFP} translates goals, decision variables, and constraints into a JSON representation, which is then used to generate Python code for an SMT solver to produce plans without task-specific examples or external critics. \textbf{NIPE} \citep{ying2023neurosymbolicinverseplanningengine} leverages LLMs as few-shot semantic parsers to generate conditional statements from spatial descriptions, guiding PDDL sampling and action model definition for Bayesian goal inference.

For real-world grounding, \textbf{MORPHeus} \citep{ye2024morpheusmultimodalonearmedrobotassisted} focuses on human-in-the-loop long-horizon planning, introducing an anomaly detection mechanism to identify potential execution errors and update corresponding PDDL files to reflect changes in the world model. \textbf{InterPret} \citep{han2024interpretinteractivepredicatelearning} uses LLMs to enable robots to learn PDDL predicates and derive action schemas through interactive language feedback from non-expert users via Python perception APIs. \citet{mahdavi2024leveragingenvironmentinteractionautomated} uses environmental interactions for evaluation and verification, starting with the LLM defining candidate PDDL problem files and domain sets, which are then refined through iterative cycles using their novel Exploration Walk (EW) method.

Instead of generating models for external planners, \textbf{AgentGen} \citep{AgentGen} uses LLMs to synthesize diverse PDDL tasks and NL descriptions for training LLM-based agents. Their work demonstrate that instruction-tuned models show significant gains in both in-domain and out-of-domain planning tasks. \citet{text2world} further fine-tune \textit{LLaMA-3.1} on this dataset, finding notable improvements in domain model generation, especially for larger models.

\begin{boxB}
    \textbf{Summary and Future Directions}: Complexities arise when coordinating the domain and respective problem. Human-in-the-loop interactions are frequently employed \citep{kelly2023there}, whereas other methods incorporate pre-processing steps---involving external tools like FastDownward and VAL (\citealp{zhou2023isrllmiterativeselfrefinedlarge}; \citealp{smirnov2024generatingconsistentpddldomains}) or custom-designed rules \citep{mahdavi2024leveragingenvironmentinteractionautomated, gestrin2024nl2planrobustllmdrivenplanning}. These linear pipelines risk cascading errors, such as the possibility of new objects in the later stages of the task, prompting new PDDL types in the domain. Future work should focus on modularity, such as enabling dynamic integration of types and predicates in later stages of generation. This would result in more adaptable and error-tolerant planning systems.
\end{boxB}

\vspace{-1mm}
\subsection{Model Editing} \label{model-editing}
\vspace{-1mm}

The use of LLMs serving more as assistive tools than fully autonomous generative solutions has shown promising applications for LLM+AP integration. Understanding LLM-editing decisions in refining specifications can support authors with greater efficiency toward an automated approach.

\citet{gragera2023exploring} investigate the limitations of LLMs in repairing unsolvable tasks caused by incorrect task specifications, assessing the effectiveness of prompting in both PDDL and NL. \citet{LLM-error-detection} conduct a comprehensive study on using LLMs, with traditional error-checking methods, to detect and correct syntactic and semantic errors in PDDL domains, demonstrating that LLMs excel at syntax correction but are less reliable with semantic inconsistencies. \citet{sikestraversing} addresses planning model failures caused by semantically equivalent but syntactically distinct state variables, a common issue when integrating information from heterogeneous sources. Their approach introduces meta-actions to bridge these mismatches and iteratively refines the model to ensure valid plan generation. \citet{caglar2024llmsfixissuesreasoning} address the challenge of modifying model spaces beyond classical planning by evaluating how effectively LLMs generate plausible model edits---especially to fix unsolvability and plan executability---to support combinatorial search and manual methods.

\begin{boxB}
    \textbf{Summary and Future Directions}: Current research shows promise in using LLMs to correct syntactic errors, but addressing semantic errors remains a significant challenge \citep{LLM-error-detection} leading to non-executable or semantically inconsistent plans. Future work should explore post-hoc correction strategies. For instance, researchers could explore strategies to analyze plan outputs, identifying semantic inconsistencies through automated metrics or human evaluation systems as grounded feedback for error-correction. 
    \end{boxB}

\vspace{-1mm}
\subsection{Model Benchmarks} \label{model-benchmarks}
\vspace{-1mm}

LLMs, with their non-deterministic output behaviors, make it challenging to assess the quality of frameworks used in planning benchmarks. This heightens the importance of robustness, especially for evaluating LLMs' ability to extract planning models \citep{behnke2024envisioning}.

\begin{boxK}
    LLM+AP benchmarks typically fall into two categories: (1) evaluating LLMs' direct planning abilities; (2) assessing the quality of planning specifications produced by LLMs. While this survey focuses on the latter, we recognize that end-to-end planning benchmarks can also support research on LLM-generated models for external planners.
\end{boxK}

\vspace{2mm}

\textit{\textbf{LLMs-as-Planner Benchmarks}}: To determine whether testing PDDL domains have been leaked to training data of LLMs, \textbf{Mystery Blocksworld} \citep{kambhampati2024llmscantplanhelp} obfuscates the classic \textbf{Blocksworld} \citep{gupta2010blocks} planning problem by altering the named types so they are semantically equivalent but syntactically nonsensical. \textbf{ALFWorld} \citep{shridhar2021alfworldaligningtextembodied} and \textbf{Household} \citet{guan2023leveragingpretrainedlargelanguage} tackles the complexities of real-world typical household environment that uses PDDL semantics to produce textual observations and high-level actions. \textbf{TravelPlanner} \citep{xie2024travelplannerbenchmarkrealworldplanning} assesses language models' abilities in planning through agent-based interactions in a travel-planning environment. \citet{zheng2024naturalplanbenchmarkingllms} extend this work with \textbf{Natural Plan}, which evaluates LLMs on realistic planning and scheduling benchmarks using APIs. \textbf{PlanBench} \citep{valmeekam2023planbench} aims to systematically evaluate LLM planning capabilities with an emphasis on cost-optimal planning and plan verification. \textbf{ACPBench} \citep{acp-bench} standardizes evaluation tasks and metrics for assessing reasoning about actions, changes (transitions), and planning—across 13 domains, on 22 SOTA language models. \textbf{AutoPlanBench} \citep{stein2024autoplanbenchautomaticallygeneratingbenchmarks} first converts PDDL planning benchmarks into NL via LLMs, and then tasks LLMs to produce a plan through various prompting techniques. \citet{benchmarking-llm} introduce a scalable benchmark suite in both PDDL and natural language to evaluate LLMs across diverse planning strategies, along with a method for translating PDDL benchmarks into natural language. \citet{puerta2025roadmap} propose a road map and benchmark to address the gap of LLM integration in Hierarchical Planning (HP).

\textit{\textbf{LLMs-as-Planning-Formalizers Benchmarks}}:
\textbf{Planetarium} \citep{zuo2024planetariumrigorousbenchmarktranslating} provides a rigorous benchmark for evaluating PDDL task/problems produced by LLMs, highlighting two key issues: (i) LLMs can produce valid code that misaligns with the original NL description, and (ii) evaluation sets often use NL descriptions too similar to the ground truth, reducing the task's challenge. The benchmark assesses LLMs' ability to generate PDDL problems across varying levels of abstraction and size. However, it currently only supports Blocksworld, Gripper, and Floor Tile domains---well-known but narrow in dataset variability. On the other hand, \textbf{Text2World} \citep{text2world} introduces an automated pipeline for domain extraction and rigorous multi-criteria metrics that address the limitations of narrow domain scope \citep{liu2023llmpempoweringlargelanguage} and indirect evaluation methods in end-to-end plan assessments. Key metrics, including executability, structural similarity, and component-wise F1 scores, are employed while exploring state-of-the-art LLMs and fine-tuning techniques. However, the reliance on executability as a gating metric excludes non-executable domains from component-wise scoring, causing minor syntax errors to skew overall quality assessments.

\begin{boxB}
        \textbf{Summary and Future Directions}: Assessing the quality of LLM-generated PDDL models \citep{zuo2024planetariumrigorousbenchmarktranslating, text2world, oswald2024largelanguagemodelsplanning} has made significant progress toward rigorous evaluation; however, the rapid leakage of training data to LLMs remains a major challenge, with \citep{text2world} reporting high contamination rates in the evaluation domains from \citep{guan2023leveragingpretrainedlargelanguage}.
        Future work should explore solutions for establishing dynamic benchmark standards for domains, actively involving the planning community in its ongoing refinement. \citet{pddl-fuse} proposes a tool for generating diverse and complex planning domains, which could serve as a foundation for such a benchmark.
    \end{boxB}

%% file: section/4-l2p.tex
\section{Language-to-Plan (L2P)}
\vspace{-1mm}
With the proliferation of related techniques to convert NL to PDDL, we are seeing an ever-increasing set of related methods. To bring them together under a single computational umbrella, and beyond just relating the work together conceptually as we have done thus far in this survey, we created a unified platform: \textbf{L2P}~\footnote{Code made publicly at: \url{https://github.com/AI-Planning/l2p}}, which re-implements landmark papers covered in this survey. This Python library is open source and has the capability of encapsulating the generalized version of the proposed ``LLM-Modulo'' framework \citep{kambhampati2024llmscantplanhelp}, which ensures soundness via iterative plan refinement with external verifiers, shifting focus from direct planning to PDDL generation with integrated verifiers and user-guided refinement through complete, planner-executable specifications.
L2P offers three major benefits: 

\begin{description} [topsep=2mm, itemsep=1mm]
    \item (i) \textbf{ Comprehensive Tool Suite}: users can easily plug in various LLMs for streamlined extraction experiments with our extensive collection of PDDL extraction and refining tools.
    \item (ii) \textbf{Modular Design}: facilitates flexible PDDL generation, allowing users to explore prompting styles and create customized pipelines.
    \item (iii) \textbf{Autonomous Capability}: supports fully autonomous end-to-end pipelines, reducing the need for manual authoring.
\end{description}

Appendix \ref{appendix:c} demonstrates examples of L2P usage. To demonstrate the flexibility of the framework, L2P re-implements some key papers covered in this survey (refer to Appendix \ref{appendix:a}). We hope to maintain the L2P framework as a repository of existing advancements in LLM model acquisition and relevant papers, ensuring that users have access to the most current research and tools.

\begin{figure}
        \begin{lstlisting}[style=python-aga]
def run_aba_alg(
    model, action_model, domain_desc, hier, prompt, max_iter: int=2
    ) -> tuple[list[Predicate], list[Action]]:
    
    actions = list(action_model.keys())
    pred_list = []
    for _ in range(max_iter):
        action_list = []
        for _, action in enumerate(actions):
            # extract action/predicates (L2P)
            pddl_action, new_preds, _, _ = (
                builder.formalize_pddl_action(
                    model=model, 
                    domain_desc=domain_desc,
                    prompt_template=prompt, 
                    action_name=action,
                    action_desc=action_model[action]['desc'],
                    types=hier["hierarchy"],
                    predicates=pred_list,
                    extract_new_preds=True
                )
            )
            pred_list.extend(new_preds)
            action_list.append(pddl_action)
    pred_list = prune_predicates(pred_list, action_list)
    return pred_list, action_list
        \end{lstlisting}
        \caption{A shortened L2P reconstruction of the 'action-by-action algorithm' \citep{guan2023leveragingpretrainedlargelanguage}, which iteratively generates PDDL actions while updating a dynamic predicate list. Output found in Figure \ref{fig:aba-output}.}
        \label{fig:l2p-example}
        \end{figure}

%% file: section/5-discussion.tex
\vspace{-1mm}
\section{Discussion}
\vspace{-2mm}
    
    \textbf{Revisiting RQ1:} While frameworks can generate parsable and solvable PDDL files, it remains uncertain if these specifications align with human goals. Simple domains like Blocksworld are easier to verify. Still, scaling complex domains requires users to understand how LLMs generate these specifications, emphasizing the need for explainable planning to yield robust, transparent, and correctable outputs \citep{zuo2024planetariumrigorousbenchmarktranslating}. Corrective feedback loops notably improve failure handling, such as resolving action precondition errors \citep{raman2024capecorrectiveactionsprecondition}, or re-planning in case of unexpected failures during plan execution \citep{joublin2023copalcorrectiveplanningrobot, raman2022planning}. Ensuring alignment with user goals involves breaking down PDDL model construction into pre-processing steps with human-in-the-loop feedback \citep{kelly2023there}. Very reminiscent of the “critics” process in the LLM-Modulo framework \citep{kambhampati2024llmscantplanhelp}, setting up a sort of external verifier checklist and using LLMs to provide feedback is demonstrated by \citet{gestrin2024nl2planrobustllmdrivenplanning} and \citet{smirnov2024generatingconsistentpddldomains}. A potential idea is analyzing the semantic correctness of plans generated and using that as feedback to refine the LLM-generated PDDL specifications \citep{Sakib_2024}. Additionally, intermediate representation (i.e. ASP, Python, JSON) that are easier for LLMs to process before converting to PDDL \citep{agarwal2024tictranslateinfercompileaccuratetext, smirnov2024generatingconsistentpddldomains} can also enhance accuracy.

    \vspace{-2mm}
    \paragraph{Revisiting RQ2:} LLMs have demonstrated that they are significantly sensitive to prompting---raising questions about whether they are better off functioning as \textit{translators} or \textit{generators}. \citet{liu2023llmpempoweringlargelanguage} demonstrate that highly explicit descriptions improve translation accuracy, while \citet{gestrin2024nl2planrobustllmdrivenplanning}  leverage minimal descriptions, relying on LLMs’ internal world knowledge to enrich outputs; however, this excess freedom often leads to inconsistent or inexecutable domain models. \citet{huang2024planningdarkllmsymbolicplanning, liu2023llmpempoweringlargelanguage, guan2023leveragingpretrainedlargelanguage} recognize that specifying a precise predicate set in NL is crucial and addresses the common problem of evaluating across different methods. The challenge of operating with minimal to no textual guidance beyond the initial task prompt underscores the importance of standardizing prompt granularity for initial generation and iterative feedback \citep{liu2024deltadecomposedefficientlongterm}. \citet{nabizada2024modelbasedworkflowautomatedgeneration} provide a promising, organized, and standardized paradigm for automatically generating PDDL descriptions that can be applied to LLMs.

%% file: section/6-conclusion.tex
\section{Conclusion}
\vspace{-2mm}
    
    Extracting planning models has long been recognized as a major barrier to the widespread adoption of planning technologies 
    \citep{vallati2020knowledge, hendler1990ai}. Even with the emergence of LLMs, this remains a persistent challenge, introducing a new suite of obstacles. In this survey, we examine nearly 80 scholarly articles that propose their frameworks and some other subsidiary works delegating model acquisition tasks to LLMs. By identifying the research distribution and gaps within these categories, we aim to provide a higher generalization from each framework’s methodologies into broader aspects for future architectures. Additionally, we hope researchers can apply these methodologies to more advanced planning languages with the support of our L2P library.

\clearpage
\section*{Limitations}

    This survey has two primary limitations. First, regarding its scope, our focus is limited to PDDL construction frameworks and related papers. Techniques remain largely unexplored in this context, and LLM capabilities in planning are still in their early stages. Due to page space constraints, we provide only a brief overview of each work rather than an exhaustive technical analysis. Additionally, our study primarily draws works published in ACL, ACM, AAAI, NeurIPS, ICAPS, COLING, CoRR, ICML, ICRA, EMNLP, and arXiv, so there is a possibility that we may have missed relevant research from other venues. Secondly, our L2P library currently supports only basic PDDL extraction tools for fully observable deterministic planning, and does not yet include tools for areas such as temporal planning. We plan to expand the library to cover a broader range of PDDL applications, aiming to further research into the challenges LLMs encounter in these areas.

\section*{Ethics Statement}
This survey does not pose any ethical issues beyond those already present in the existing literature on planning model construction via LLMs. As with any sufficiently advanced technology, there is an opportunity for misuse of the proposed L2P library (e.g., extracting actionable planning models for unethical domains). However, we view this as a pervasive issue with all of the existing methods that aim to extract planning theories from natural language.

%% file: section/appendix.tex
\onecolumn
\newpage
\appendix

\section{Paper Overview}
\label{appendix:a}
Below is a quick summary of core model generation frameworks in this survey. Papers are sorted by planning specification coverage, followed by publication year, and then alphabetically by author.

\begin{figure}[htbp]
  \centering
  \scriptsize
  \renewcommand{\arraystretch}{1.3}
  \begin{tabular}{c@{ }c@{ } | c@{  }c@{ }c@{  }c | c c c}
    \hline
    \multicolumn{2}{c|}{\textbf{WORKS}} & \multicolumn{4}{c|}{\textbf{PDDL SPEC.}} & \multicolumn{3}{c}{\textbf{GUIDANCE}} \\
    \textbf{Paper} & \textbf{Framework} & \textbf{Init.} & \textbf{Goal} & \textbf{Preds.} & \textbf{Act.} & \textbf{Prompt} & \textbf{Feedback} & \textbf{Human Int.} \\
    \hline

    $^\star$\citep{gestrin2024nl2planrobustllmdrivenplanning} & NL2PLAN & $\checkmark$ & $\checkmark$ & $\checkmark$ & $\checkmark$ & Heavy (Few-shot+CoT) & Custom validator + LLM & Optional \\
    \hline

    \citep{liu2024deltadecomposedefficientlongterm} & DELTA & $\checkmark$ & $\checkmark$ & $\checkmark$ & $\checkmark$ & Medium (Few-shot) & None & None \\
    \hline

    \citep{mahdavi2024leveragingenvironmentinteractionautomated} & LLM+EW & $\checkmark$ & $\checkmark$ & $\checkmark$ & $\checkmark$ & Light & Env. (EW Metric) + LLM & None \\
    \hline

    \citep{smirnov2024generatingconsistentpddldomains} & --- & $\checkmark$ & $\checkmark$ & $\checkmark$ & $\checkmark$ & Light (JSON interm.) & Custom validator + LLM & Optional \\
    \hline

    \citep{Sakib_2024} & --- & $\checkmark$ & $\checkmark$ & $\checkmark$ & $\checkmark$ & Medium (Few-shot) & None & None \\
    \hline

    \citep{ye2024morpheusmultimodalonearmedrobotassisted} & MORPHeus & $\checkmark$ & $\checkmark$ & $\checkmark$ & $\checkmark$ & Heavy (Few-shot+CoT) & NL Human feedback & True \\
    \hline

    \citep{kelly2023there} & TABA & $\checkmark$ & $\checkmark$ & $\checkmark$ & $\checkmark$ & Light (One-shot) & Glaive validator + LLM & Optional \\
    \hline

    \citep{ying2023neurosymbolicinverseplanningengine} & NIPE & $\checkmark$ & $\checkmark$ & $\checkmark$ & $\checkmark$ & Medium (Few-shot) & None & None \\
    \hline
    
    \citep{zhou2023isrllmiterativeselfrefinedlarge} & ISR-LLM & $\checkmark$ & $\checkmark$ & $\checkmark$ & $\checkmark$ & Medium (Few-shot) & None & None \\
    \hline

    \citep{han2024interpretinteractivepredicatelearning} & InterPret & $\times$ & $\checkmark$ & $\checkmark$ & $\checkmark$ & Light (CoT) & NL Human feedback & True \\
    \hline

    \citep{liu2024humanawarenessrobottask} & --- & $\times$ & $\checkmark$ & $\checkmark$ & $\checkmark$ & Medium (unknown) & Scene graph & True \\
    \hline

    $^\star$\citep{text2world} & TEXT2WORLD & $\times$ & $\times$ & $\checkmark$ & $\checkmark$ & Heavy (Few-shot+Exp.) & Tarski validator + LLM & None \\
    \hline

    \citep{sinha2024leveraging} & --- & $\times$ & $\times$ & $\checkmark$ & $\checkmark$ & Heavy (Few-shot+Exp.) & None & None \\
    \hline

    $^\star$\citep{sikes2024creating} & --- & $\times$ & $\times$ & $\checkmark$ & $\checkmark$ & Medium (Few-shot) & None & None \\
    \hline

    $^\star$\citep{guan2023leveragingpretrainedlargelanguage} & LLM+DM & $\times$ & $\times$ & $\checkmark$ & $\checkmark$ & Heavy (Few-shot+CoT+Exp.) & Custom validator + LLM & None \\
    \hline

    \citep{chen2024languageaugmentedsymbolicplanneropenworld} & LASP & $\times$ & $\times$ & $\times$ & $\checkmark$ & Medium (High Exp.) & LLM & None \\
    \hline

    \citep{huang2024planningdarkllmsymbolicplanning} & --- & $\times$ & $\times$ & $\times$ & $\checkmark$ & Medium (Few-shot) & VSCode-PDDL validator & None \\
    \hline

    \citep{liulearning} & BLADE & $\times$ & $\times$ & $\times$ & $\checkmark$ & Heavy & Human demonstration & True \\
    \hline

    \citep{oates2024using} & CLLaMP & $\times$ & $\times$ & $\times$ & $\checkmark$ & Medium (Few-shot) & None & None \\
    \hline

    $^\star$\citep{oswald2024largelanguagemodelsplanning} & NL2PDDL & $\times$ & $\times$ & $\times$ & $\checkmark$ & Medium (Few-shot) & None & None \\
    \hline

    \citep{wong2023learningadaptiveplanningrepresentations} & ADA & $\times$ & $\times$ & $\times$ & $\checkmark$ & Medium (Few-shot+Exp.) & None & None \\
    \hline

    $^\star$\citep{zhang2024proc2pddlopendomainplanningrepresentations} & PROC2PDDL & $\times$ & $\times$ & $\times$ & $\checkmark$ & Heavy (Few-shot) & None & None \\
    \hline

    \citep{athalye2024predicate} & pix2pred & $\times$ & $\times$ & $\checkmark$ & $\times$ & Light & Candidate filtering & None \\
    \hline

    \citep{IALP} & IALP & $\checkmark$ & $\checkmark$ & $\times$ & $\times$ & Medium (Few-shot+VLM) & VLM & None \\
    \hline

    \citep{agarwal2024tictranslateinfercompileaccuratetext} & TIC & $\checkmark$ & $\checkmark$ & $\times$ & $\times$ & Light (Zero/Few-shot) & None & None \\
    \hline

    \citep{delarosa2024trippaltravelplanningguarantees} & TRIP-PAL & $\checkmark$ & $\checkmark$ & $\times$ & $\times$ & Light & None & None \\
    \hline

    \citep{CaSTL} & CaSTL & $\checkmark$ & $\checkmark$ & $\times$ & $\times$ & Medium (Exp.) & Python validator + LLM & None \\
    \hline

    \citep{kalland2024enabling} & SemReBot2 & $\checkmark$ & $\checkmark$ & $\times$ & $\times$ & Medium (Few-shot+Exp.) & None & None \\
    \hline

    \citep{fast-accurate-llm} & --- & $\checkmark$ & $\checkmark$ & $\times$ & $\times$ & Medium (Few-shot) & None & None \\
    \hline

    \citep{lee2024planning} & PlanAID & $\checkmark$ & $\checkmark$ & $\times$ & $\times$ & Light (Single-shot) & Human deviation & True \\
    \hline

    \citep{zhang2024pddlegoiterativeplanningtextual} & PDDLEGO & $\checkmark$ & $\checkmark$ & $\times$ & $\times$ & Medium (Few-shot+Exp.) & LLM (PDDL-edit) & None \\
    \hline

    \citep{zhang2024lamma} & LaMMA-P & $\checkmark$ & $\checkmark$ & $\times$ & $\times$ & Medium (Few-shot) & FastDownward & None \\
    \hline

    $^\star$\citep{liu2023llmpempoweringlargelanguage} & LLM+P & $\checkmark$ & $\checkmark$ & $\times$ & $\times$ & Medium (Few-shot) & None & None \\
    \hline

    \citep{ViLaIn} & ViLaIn & $\checkmark$ & $\checkmark$ & $\times$ & $\times$ & Medium (Few-shot+VLM) & FastDownward & None \\
    \hline

    \citep{Birr_2024} & Auto-GPT+P & $\times$ & $\checkmark$ & $\times$ & $\times$ & Light & PDDL validator + Prolog & True \\
    \hline

    \citep{grover2024demonstration} & --- & $\times$ & $\checkmark$ & $\times$ & $\times$ & Medium (Few-shot) & AI2Thor Env. & None \\
    \hline

    \citep{izquierdo2024plancollabnl} & PlanCollabNL & $\times$ & $\checkmark$ & $\times$ & $\times$ & Light & LLM & None \\
    \hline

    \citep{SafePlanner} & SafePlanner & $\times$ & $\checkmark$ & $\times$ & $\times$ & Medium (Exp.) & None & None \\
    \hline

    \citep{paulius2024bootstrappingobjectlevelplanninglarge} & OLP & $\times$ & $\checkmark$ & $\times$ & $\times$ & Light & None & None \\
    \hline

    \citep{singh2024anticipatecollabdatadriven} & DaTAPlan & $\times$ & $\checkmark$ & $\times$ & $\times$ & Medium (Few-shot/CoT+Exp.) & Human deviation & True \\
    \hline

    \citep{singh2024twostepmultiagenttaskplanning} & TwoStep & $\times$ & $\checkmark$ & $\times$ & $\times$ & Medium (Few-shot) & LLM & None \\
    \hline

    \citep{dagan2023dynamicplanningllm} & LLM+DP & $\times$ & $\checkmark$ & $\times$ & $\times$ & Medium (Few-shot+Exp.) & Alfworld Env. & None \\
    \hline

    \citep{lyu2023faithfulchainofthoughtreasoning} & Faithful CoT & $\times$ & $\checkmark$ & $\times$ & $\times$ & Light (CoT) & None & None \\
    \hline

    \citep{xie2023translatingnaturallanguageplanning} & --- & $\times$ & $\checkmark$ & $\times$ & $\times$ & Medium (Few-shot) & None & None \\
    \hline

    $^\star$\citep{collins2022structuredflexiblerobustbenchmarking} & P+S & $\times$ & $\checkmark$ & $\times$ & $\times$ & Medium (Few-shot) & None & None \\
    \hline
    
  \end{tabular}
  \caption{\textit{Exp.} = Explicit PDDL info. \textit{Feedback/Human Intervention} provided at the level of the LLM-generated PDDL spec. itself. $^{\star}$Papers reconstructed by L2P.}
\end{figure}

\newpage

\section{Additional Information on PDDL}
\label{appendix:b}

\subsection{Why PDDL?} \label{appendix:why-pddl}
PDDL, as a \textbf{declarative} language, differs fundamentally from \textbf{imperative} languages like Python in how problems are expressed and solved. Declarative programming specifies \textit{what} needs to be achieved rather than \textit{how} to achieve it, leaving execution to an external planner. In contrast, imperative programming defines explicit step-by-step instructions, requiring precise control over execution flow. While LLMs struggle with logical reasoning in imperative programming due to its sequential dependencies, they perform well with declarative representations like PDDL. Using LLMs for PDDL model construction is \textbf{not traditional code generation} but rather knowledge structuring: organizing states, actions, and constraints into a formalized model. Instead of writing executable logic, LLMs assist in mapping natural language descriptions to structured symbolic representations.

\subsection{PDDL Example}
Automated Planning is a specialized field within AI that can be challenging for those unfamiliar with its principles. A key tool in classical planning is the Planning Domain Definition Language (PDDL), which models planning problems and domains. We illustrate its concepts with the Blocksworld problem, where blocks must be stacked in a specific order using actions like picking up, unstacking, and placing blocks, all while respecting constraints like moving only one block at a time or not disturbing stacked blocks.

Demonstrated below is the Blocksworld PDDL domain file \(\mathbb{DF}\):

\begin{lstlisting}[style=pddlstyle]
(define (domain blocksworld)
    (:requirements :strips)
    (:predicates (clear ?x) (on-table ?x) (arm-empty) (holding ?x) (on ?x ?y))
    (:action pickup
        :parameters (?ob)
        :precondition (and (clear ?ob) (on-table ?ob) (arm-empty))
        :effect (and (holding ?ob) (not (clear ?ob)) (not (on-table ?ob)) (not (arm-empty)))
    ))
\end{lstlisting}

\textbf{Predicates} define relationships or properties that can be true or false, such as \texttt{(on ?x ?y)} for block \texttt{?x} on \texttt{?y}, \texttt{(ontable ?x)} for \texttt{?x} on the table, \texttt{(clear ?x)} for \texttt{?x} having nothing on top, and \texttt{(holding ?x)} for the robot holding \texttt{?x}. \textbf{Actions} describe possible state changes. For instance, \texttt{(pick-up)} contains the \textit{parameter(s)} block \texttt{?ob}, \textit{preconditions} requiring \texttt{(clear ?ob)} and \texttt{(ontable ?ob)}, and \textit{effects} updating the state to reflect the robot holding \texttt{?ob}, which is no longer on the table and clear.

The following is the corresponding PDDL problem/task file \(\mathbb{PF}\):

\begin{lstlisting}[style=pddlstyle]
(define (problem blocksworld-problem) 
   (:domain blocksworld) 
   (:objects A B C) ; Blocks 
   (:init (ontable A) (ontable B) (on C A) (clear B) (clear C)) ; Initial state 
   (:goal (and (on A B) (on B C)))) ; Goal state
\end{lstlisting}

\textbf{Objects} represent the entities involved, such as blocks A, B, and C. The \textbf{initial state} defines the starting arrangement, where blocks A and B are on the table, block C is on A, and both B and C are clear. The \textbf{goal state} specifies the desired configuration, where block A is stacked on B, and block B is stacked on C.

Given the above PDDL domain and problem, a classical planner might generate the following plan:

\begin{lstlisting}[style=pddlstyle]
Unstack C from A 
Put C on table 
Pick up A 
Stack A on B 
Pick up B 
Stack B on C 
\end{lstlisting}

\section{(L2P) Framework}
\label{appendix:c}

\subsection{General Library Overview}
L2P provides a comprehensive suite of tools for PDDL model creation and validation. The Builder classes enable users to prompt the LLM to generate essential components of a PDDL domain, such as types, predicates, and actions, along with their parameters, preconditions, and effects. It also supports task specification, including objects, initial, and goal states that correspond to the given domain. L2P features a customized Feedback Builder class that incorporates both LLM-generated feedback and human input, or a combination of the two. Additionally, the library includes a syntax validation tool that detects common PDDL syntax errors, using this feedback to improve the accuracy of the generated models---as illustrated in Figure \ref{fig:formal-specification}, which shows how LLM feedback can be used to refine a PDDL problem specification.

\subsection{Paper Reconstructions}
    L2P can recreate and encompass previous frameworks for converting natural language to PDDL, serving as a comprehensive foundation that integrates past approaches. L2P contains multiple paper reconstructions as examples of how existing methods can be implemented and compared within a unified system, demonstrating L2P's flexibility and effectiveness in standardizing diverse NL-to-PDDL techniques. An example of \citet{guan2023leveragingpretrainedlargelanguage} ``action-by-action'' algorithm can be found in Figure \ref{fig:l2p-example}; action and predicate output can be found in Figure \ref{fig:aba-output}.

\subsection{L2P Usage Example}

Below are example usages using our L2P library. Full documentation can be found on our website.

    \begin{figure*}[h!]
        \scriptsize
        \begin{lstlisting}[style=python]   
    import os
    from l2p.llm.openai import OPENAI
    from l2p.utils import load_file
    from l2p.domain_builder import DomainBuilder
    
    domain_builder = DomainBuilder()
    
    api_key = os.environ.get('OPENAI_API_KEY')
    llm = OPENAI(model="gpt-4o-mini", api_key=api_key)
    
    # retrieve prompt information
    base_path='tests/usage/prompts/domain/'
    domain_desc = load_file(f'{base_path}blocksworld_domain.txt')
    predicates_prompt = load_file(f'{base_path}formalize_predicates.txt')
    types = load_file(f'{base_path}types.json')
    action = load_file(f'{base_path}action.json')
    
    # extract predicates via LLM
    predicates, llm_output, validation_info = domain_builder.formalize_predicates(
        model=llm,
        domain_desc=domain_desc,
        prompt_template=predicates_prompt,
        types=types
        )
    
    # format key info into PDDL strings
    predicate_str = "\n".join([pred["raw"].replace(":", " ; ") for pred in predicates])
    
    print(f"###OUTPUT\n{predicate_str}")
    -------------------------------------------------------------------------------------
    
    ### OUTPUT
    - (holding ?a - arm ?b - block) ;  true if the arm ?a is currently holding the block ?b
    - (on_table ?b - block) ;  true if the block ?b is on the table
    - (clear ?b - block) ;  true if the block ?b is clear (no block on top of it)
    - (on_top ?b1 - block ?b2 - block) ;  true if the block ?b1 is on top of the block ?b2
        \end{lstlisting}
        \caption{L2P usage - generating simple PDDL predicates}
        \label{fig:l2p-usage}
        \end{figure*}

\begin{figure*}[h!]
    \small
    \begin{lstlisting}[style=python]       
    import os
    from l2p.utils.pddl_types import Predicate
    from l2p.task_builder import TaskBuilder
    
    task_builder = TaskBuilder() # initialize task builder class
    api_key = os.environ.get('OPENAI_API_KEY')
    llm = OPENAI(model="gpt-4o-mini", api_key=api_key)
    
    # load in assumptions
    problem_desc = load_file(r'tests/usage/prompts/problem/blocksworld_problem.txt')
    task_prompt = load_file(r'tests/usage/prompts/problem/formalize_task.txt')
    types = load_file(r'tests/usage/prompts/domain/types.json')
    predicates_json = load_file(r'tests/usage/prompts/domain/predicates.json')
    predicates: List[Predicate] = [Predicate(**item) for item in predicates_json]
    
    # extract PDDL task specifications via LLM
    objects, init, goal, llm_response, validation_info = task_builder.formalize_task(
        model=llm,
        problem_desc=problem_desc,
        prompt_template=task_prompt,
        types=types,
        predicates=predicates
        )
    
    # generate task file
    pddl_problem = task_builder.generate_task(
        domain_name="blocksworld",
        problem_name="blocksworld_problem",
        objects=objects,
        initial=init,
        goal=goal)
    
    print(f"### LLM OUTPUT:\n {pddl_problem}")
    -------------------------------------------------------------------------------------
    ### LLM OUTPUT:
    (define
       (problem blocksworld_problem)
       (:domain blocksworld)
    
       (:objects 
          arm1 - arm
          blue_block - block
          red_block - block
          yellow_block - block
          green_block - block
       )
    
       (:init
          (on_top blue_block red_block)
          (on_top red_block yellow_block)
          (on_table yellow_block)
          (on_table green_block)
          (clear blue_block)
          (clear green_block)
       )
    
       (:goal
          (and 
             (on_top red_block green_block)
          )
       )
    )
    \end{lstlisting}
    \caption{L2P usage - generating simple PDDL task specification}
    \end{figure*}

\begin{figure*}[h!]
    \scriptsize
    \begin{lstlisting}[style=python]
    import os  
    from l2p.feedback_builder import FeedbackBuilder

    feedback_builder = FeedbackBuilder()
    api_key = os.environ.get('OPENAI_API_KEY')
    llm = OPENAI(model="gpt-4o-mini", api_key=api_key)
    
    problem_desc = load_file(r'tests/usage/prompts/problem/blocksworld_problem.txt')
    types = load_file(r'tests/usage/prompts/domain/types.json')
    feedback_template = load_file(r'tests/usage/prompts/problem/feedback.txt')
    predicates_json = load_file(r'tests/usage/prompts/domain/predicates.json')
    predicates: list[Predicate] = [Predicate(**item) for item in predicates_json]
    llm_response = load_file(r'tests/usage/prompts/domain/llm_output_task.txt')
    
    fb_pass, feedback_response = feedback_builder.task_feedback(
        model=llm,
        problem_desc=problem_desc,
        llm_output=llm_response,
        feedback_template=feedback_template,
        feedback_type="llm",
        predicates=predicates,
        types=types)
    
    print("[FEEDBACK]\n", feedback_response)
    -------------------------------------------------------------------------------------
    [FEEDBACK]
    ### JUDGMENT
    ```
    My feedback on the provided PDDL problem file is as follows:
    
    1. Are any necessary objects missing?
       All necessary blocks are included. Therefore: No.
    
    2. Are any unnecessary objects included?
       All objects are relevant to the problem. Hence: No.
    
    3. Are any objects defined with the wrong type?
       All objects are correctly defined as "object". Therefore: No.
    
    4. Are any unnecessary or incorrect predicates declared?
       All predicates used are relevant and correctly applied. Thus: No.
    
    5. Are any needed or expected predicates missing from the initial state?
       The initial state is missing the predicate for the red block being not clear, as it is covered by the blue block. This should be added:
       - (clear red_block) should be false, but it is not explicitly stated. Hence: Yes.
    
    6. Is anything missing from the goal state?
       The goal state is correctly defined as having the red block on top of the green block. So: No.
    
    7. Is anything unnecessary included in the goal description?
       The goal description is concise and only includes what is necessary. Therefore: No.
    
    8. Should any predicate be used in a symmetrical manner?
       The predicates used do not require symmetry in this context. Hence: No.
    
    In summary, the main issue is the missing predicate regarding the clarity of the red block. It should be explicitly stated that the red block is not clear due to the blue block being on top of it. 
    
    To improve the initial state, you should add:
    - (clear red_block) should be false, or alternatively, you can add:
    - (not (clear red_block)) to indicate that the red block is not clear.
    ```
    \end{lstlisting}
    \caption{L2P usage - generating LLM-feedback on task specification}
    \label{fig:formal-specification}
    \end{figure*}

\begin{figure}[t]
    \centering
    \begin{boxK}
    \scriptsize
    \begin{verbatim}
## PREDICATES
{'name': 'truck-at', 
    'desc': 'true if the truck ?t is currently at location ?l', 
    'raw': '(truck-at ?t - truck ?l - location): true if the truck ?t is currently at location ?l', 
    'params': OrderedDict([('?t', 'truck'), ('?l', 'location')]), 
    'clean': '(truck-at ?t - truck ?l - location): true if the truck ?t is currently at location ?l'}
{'name': 'package-at', 
    'desc': 'true if the package ?p is currently at location ?l', 
    'raw': '(package-at ?p - package ?l - location): true if the package ?p is currently at location ?l', 
    'params': OrderedDict([('?p', 'package'), ('?l', 'location')]), 
    'clean': '(package-at ?p - package ?l - location): true if the package ?p is currently at location ?l'}
{'name': 'truck-holding', 
    'desc': 'true if the truck ?t is currently holding the package ?p', 
    'raw': '(truck-holding ?t - truck ?p - package): true if the truck ?t is currently holding the package ?p', 
    'params': OrderedDict([('?t', 'truck'), ('?p', 'package')]), 
    'clean': '(truck-holding ?t - truck ?p - package): true if the truck ?t is currently holding the package ?p'}
{'name': 'truck-has-space', 
    'desc': 'true if the truck ?t has space to load more packages', 
    'raw': '(truck-has-space ?t - truck): true if the truck ?t has space to load more packages', 
    'params': OrderedDict([('?t', 'truck')]), 
    'clean': '(truck-has-space ?t - truck): true if the truck ?t has space to load more packages'}
{'name': 'plane-at', 
    'desc': 'true if the airplane ?a is located at location ?l', 
    'raw': '(plane-at ?a - plane ?l - location): true if the airplane ?a is located at location ?l', 
    'params': OrderedDict([('?a', 'plane'), ('?l', 'location')]), 
    'clean': '(plane-at ?a - plane ?l - location): true if the airplane ?a is located at location ?l'}
{'name': 'plane-holding', 
    'desc': 'true if the airplane ?a is currently holding the package ?p', 
    'raw': '(plane-holding ?a - plane ?p - package): true if the airplane ?a is currently holding package ?p', 
    'params': OrderedDict([('?a', 'plane'), ('?p', 'package')]), 
    'clean': '(plane-holding ?a - plane ?p - package): true if the airplane ?a is currently holding package ?p'}
{'name': 'connected-locations', 
    'desc': 'true if location ?l1 is directly connected to location ?l2 in city ?c', 
    'raw': '(connected-locations ?l1 - location ?l2 - location ?c - city): ?l1 is connected to ?l2 in city ?c', 
    'params': OrderedDict([('?l1', 'location'), ('?l2', 'location'), ('?c', 'city')]), 
    'clean': '(connected-locations ?l1 - location ?l2 - location ?c - city): ?l1 is connected to ?l2 in city ?c'}

## ACTIONS
{'name': 'load_truck', 'parameters': OrderedDict([('?p', 'package'), ('?t', 'truck'), ('?l', 'location')]), 
    'preconditions': '(and\n    (truck-at ?t ?l)\n    (package-at ?p ?l)\n    (truck-has-space ?t)\n)', 
    'effects': '(and\n    (not (package-at ?p ?l))\n    (truck-holding ?t ?p)\n)'}
{'name': 'unload_truck', 'parameters': OrderedDict([('?p', 'package'), ('?t', 'truck'), ('?l', 'location')]), 
    'preconditions': '(and\n    (truck-at ?t ?l)\n    (truck-holding ?t ?p)\n)', 
    'effects': '(and\n    (not (truck-holding ?t ?p))\n    (package-at ?p ?l)\n)'}
{'name': 'load_airplane', 'parameters': OrderedDict([('?p', 'package'), ('?a', 'plane')]), 
    'preconditions': '(and\n    (package-at ?p ?l)\n    (plane-at ?a ?l)\n)', 
    'effects': '(and\n    (not (package-at ?p ?l))\n    (plane-holding ?a ?p)\n)'}
{'name': 'unload_airplane', 'parameters': OrderedDict([('?p', 'package'), ('?a', 'plane'), ('?l', 'location')]), 
    'preconditions': '(and\n    (plane-at ?a ?l)\n    (plane-holding ?a ?p)\n)', 
    'effects': '(and\n    (not (plane-holding ?a ?p))\n    (package-at ?p ?l)\n)'}
{'name': 'drive_truck', 
    'parameters': OrderedDict([('?t', 'truck'), ('?l1', 'location'), ('?l2', 'location'), ('?c', 'city')]), 
    'preconditions': '(and\n    (truck-at ?t ?l1)\n    (connected-locations ?l1 ?l2 ?c)\n)', 
    'effects': '(and\n    (not (truck-at ?t ?l1))\n    (truck-at ?t ?l2)\n)'}
{'name': 'fly_airplane', 
    'parameters': OrderedDict([('?a', 'plane'), ('?l1', 'location'), ('?l2', 'location'), ('?c', 'city')]), 
    'preconditions': '(and\n    (plane-at ?a ?l1)\n    (connected-locations ?l1 ?l2 ?c)\n)', 
    'effects': '(and\n    (not (plane-at ?a ?l1))\n    (plane-at ?a ?l2)\n)'}
\end{verbatim}
    \end{boxK}
    \caption{L2P formatted predicate and actions outputted from LLM (gpt-4o-mini) via `action-by-action` algorithm \citep{guan2023leveragingpretrainedlargelanguage} on Logistics domain.}
    \label{fig:aba-output}
\end{figure}